\documentclass[11pt]{article}
\usepackage[preprint]{acl}
\usepackage{times}
\usepackage[T1]{fontenc}
\usepackage{microtype}
\usepackage{amsmath,amssymb}
\usepackage{booktabs}
\usepackage{graphicx}
\usepackage[nameinlink,noabbrev]{cleveref}
\usepackage{enumitem}
\newcommand{\meansd}[2]{\ensuremath{#1_{\pm #2}}}

\title{Training LLMs for Multi-Agent Collaboration with Verifiable Information Rewards}
\author{
  Keru Chen \\
  School of ECEE \\
  Arizona State University \\
  \texttt{kchen234@asu.edu} 
  \And
  Sen Lin \\
  Computer Science Department \\
  University of Houston \\
  \texttt{slin50@central.uh.edu} 
  \AND
  Yingbin Liang \\
  Department of ECE \\
  The Ohio State University \\
  \texttt{liang.889@osu.edu} 
  \And
  Nathaniel D. Bastian \\
  Whiting School of Engineering \\
  Johns Hopkins University \\
  \texttt{ndbastian@jhu.edu} 
  \And
  Shaofeng Zou\thanks{\ \ Corresponding author.} \\
  School of ECEE \\
  Arizona State University \\
  \texttt{zou@asu.edu}
}

\begin{document}
\maketitle

\begin{abstract}
Effective collaboration among large language model (LLM) agents in multi-agent systems requires sharing relevant evidence, requesting missing information, and incorporating teammates’ contributions into decisions. Training these behaviors is challenging because final task outcomes provide limited feedback on the value of individual messages. We introduce CoSIR (Collaboration via Surprisal-based Information Rewards), a post-training approach that uses a simple, verifiable environment to teach collaboration behaviors that generalize to more complex, unseen tasks. We propose a Bayesian evaluator that scores shared information by its reduction in team surprisal, defined as the sum of the negative log probabilities assigned to the correct option given each agent’s available evidence. Requests are scored by the potential reduction in the same quantity if the requested information were truthfully provided. Separate rewards encourage agents to use the received information in their private choices and calculations.
%
% We propose a Bayesian evaluator that quantifies the value of shared information through reductions in team surprisal: the sum of agents’ negative log probabilities of the correct option, and evaluates requests through hypothetical truthful replies. 
% In multi-agent systems, information needed to solve a task is often distributed across agents. However, large language model (LLM) agents may omit relevant facts or fail to use the information they receive. Training these behaviors requires rewards that identify how individual messages help the team make a correct decision. We introduce CoSIR (Collaboration via Surprisal-based Information Rewards), which trains collaboration in a toy environment where agents see different parts of a rating table and communicate to choose the option with the highest total rating. The table lets us identify which ratings agents share or request. A Bayesian evaluator uses each agent's available information to calculate the probability that each option is best. CoSIR rewards statements that increase the evaluator's probability of the correct option and requests whose truthful answers would provide such information. These rewards use reductions in team surprisal, the sum of the negative log probabilities of the correct option across agents. Separate rewards train agents to use the received ratings in their private choices and calculations. 
We post-train Qwen3-4B, Qwen3-8B, and Gemma-3-4B-it through self-play and evaluate transfer to seven unseen natural-language collaboration tasks. CoSIR achieves the highest aggregate score, while its performance in general tasks remains competitive. Behavioral analyses show increased disclosure of private evidence and fewer judged failures to use teammates’ input. %These results suggest that verifiable synthetic environments can provide useful training signals for collaboration beyond the training tasks.
% We post-train Qwen3-4B, Qwen3-8B, and Gemma-3-4B-it through self-play reinforcement learning, with numerical rewards computed from the table and information records rather than a learned reward judge. 
% On seven unseen natural-language collaboration tasks, each trained model has higher mean scores than its starting checkpoint on every task and the highest sum of task scores among the compared methods. Scores on mathematical reasoning, general knowledge, and instruction-following benchmarks remain close to those of the starting models.
\end{abstract}

% Main figure placed here to float to the top of page 2.
\begin{figure*}[!t]
\centering
\includegraphics[width=\textwidth]{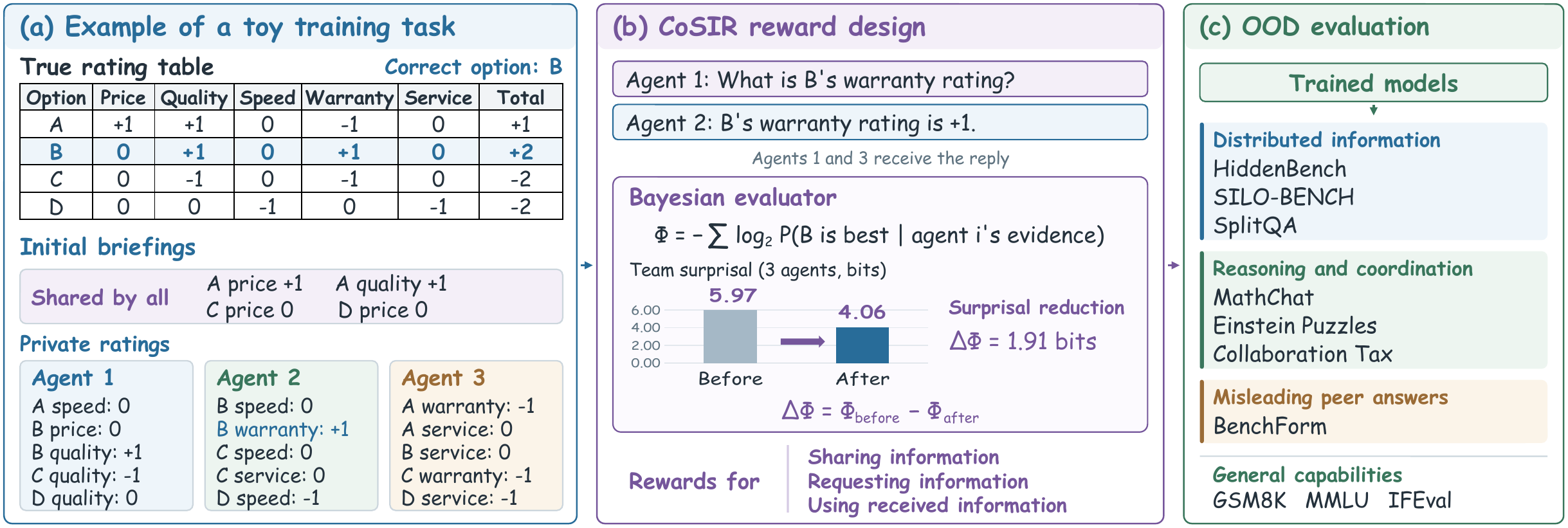}
\caption{Overview of CoSIR. Agents receive different parts of a rating table and exchange messages before choosing an option. A Bayesian evaluator scores shared ratings by how they change the probability of the correct option and scores requests by the potential value of a truthful answer. Private-answer rewards assess option choices and calculations using the ratings available to each agent. Self-play updates a policy shared by all agents, which is then evaluated on seven unseen collaboration tasks. The table and briefings illustrate one training instance.}
\label{fig:training-overview}
\end{figure*}

\section{Introduction}
\label{sec:introduction}
Effective collaboration requires teams to turn information held by individual members into evidence that informs a shared decision. Studies of human collective intelligence suggest that team performance depends on more than members' individual abilities \citep{woolley2010collective}. 
For large language model (LLM) agents, however, access to sufficient information at the team level does not guarantee successful collaboration: a team with distributed evidence can perform worse than a single agent given the same evidence in full \citep{li2026hiddenbench}.
Documented failures include withholding useful findings, proceeding without clarification, and ignoring teammates' input \citep{cemri2025fail}. Agents may also abandon correct answers in favor of an incorrect majority \citep{weng2025conformity}. These observations highlight the need for agents to communicate, seek missing information, and leverage teammate input.

LLM post-training largely emphasizes instruction following \citep{ouyang2022training,chen2026hipo} and multi-turn task completion \citep{wang2025ragen}, but these objectives do not explicitly distinguish how individual messages contribute to a team's decision. 
Recent work has begun to train collaboration more directly. For example, MAPoRL rewards correct answers and corrective or persuasive discussion \citep{park2025maporl}. We focus on an essential question: how can agents be rewarded for providing or requesting information that helps teammates identify the correct answer? Such rewards require assessing the value of a message relative to what its recipients already know. This is difficult in open-ended natural-language tasks, where the relevant facts, their distribution across agents, and their contributions to the final decision are not readily verifiable.

% A central challenge, however, is that in open-ended natural-language tasks, it is difficult to quantify how much a particular statement or request helps the team reach the correct answer. This makes it hard to provide direct training rewards for these behaviors. 
% Yet many complex collaboration tasks share a basic challenge: agents hold different pieces of information and must decide what to share, what to ask, and how to use their teammates' replies. We therefore isolate these decisions in a simpler environment that preserves the need to exchange and combine information while making the value of each exchange computable. By generating diverse tasks and calculating rewards automatically, we can provide fine-grained training feedback at scale rather than strategies specific to the training task.
We introduce CoSIR (Collaboration via Surprisal-based Information Rewards), an approach that trains collaboration in simplified, verifiable environments while preserving the information dependencies underlying more complex tasks. The key is to retain the need to share relevant evidence, request missing information, and integrate teammates' contributions, while making the value of these behaviors computable. By varying the available evidence and its distribution across agents, such environments provide diverse opportunities to learn these decisions with fine-grained, automatically computed feedback. The goal is to develop transferable collaboration behaviors in a controlled setting and evaluate whether they generalize to more complex, unseen natural-language tasks.

To instantiate this approach, we construct a task in which agents jointly select the option with the highest total rating across several attributes. Each agent observes only part of a rating table, creating a need to share known ratings and request missing information from teammates. Because reported and requested ratings correspond to specific table entries, we can identify the information exchanged, verify reported values, and track the evidence available to each agent.
%The discrete rating space further enables exact computation, under a specified probabilistic model, of how likely each option is to be optimal given that evidence. This provides the basis for evaluating how individual exchanges change the evidence supporting the correct decision.
% Because each rating takes only three possible values, we can enumerate the possible values of missing ratings and determine which option would be best in each possible table.

% To provide direct rewards for collaboration, we construct a simple training task in which agents choose the option with the highest total rating across several attributes. Each agent receives an initial briefing containing only part of the rating table and must share known ratings or request missing ones from teammates. Unlike open-ended tasks, the information exchanged in this setting can be identified as ratings for specific options and attributes. We can therefore identify which ratings agents report or request and track which reports their teammates receive. Because each rating takes only three possible values, we can enumerate the possible values of missing ratings and determine which option would be best under each possible table.

We use a Bayesian evaluator to estimate the probability that each option is optimal given an agent's initial briefing and received reports.  The complete table determines the true best option, but its hidden ratings are not used to calculate these probabilities. We define \textbf{team surprisal} as the sum, across agents, of the negative log probabilities assigned by the evaluator to the correct option. A shared rating is valued by the reduction in team surprisal it produces, while a request is valued by the potential reduction if the requested rating were truthfully provided. These quantities provide fine-grained training signals for sharing and requesting useful information, with numerical rewards computed from the table and information records rather than assigned by a learned reward judge.
% Reductions in team surprisal provide rewards for sharing and requesting information without relying on a learned reward judge. Our goal is not to train models specifically for this rating-table task, but to use its computable rewards to teach general collaboration behaviors that transfer to more complex natural-language tasks.

To encourage agents to use the information they receive, we complement communication rewards with supervision of private answers, including option choices and calculated totals.
%
% Sharing and requesting information are not enough if agents cannot use what they receive. We therefore also evaluate private answers containing an option choice and calculated totals. 
The choice score uses the probability that the selected option is best given the agent's available information, while the calculation score checks whether the agent correctly combines its known ratings. We use these rewards to post-train Qwen3-4B, Qwen3-8B \citep{yang2025qwen3}, and Gemma-3-4B-it \citep{gemmateam2025gemma3} through self-play reinforcement learning.

We test the transfer ability on seven unseen tasks: HiddenBench \citep{li2026hiddenbench}, SILO-BENCH \citep{zhang2026silo}, SplitQA, MathChat \citep{wu2023mathchat}, BenchForm \citep{weng2025conformity}, Einstein Puzzles \citep{peng2025communication}, and Collaboration Tax \citep{sun2026collaboration}. We introduce SplitQA to test whether models trained to choose among options can combine facts held by different agents to produce free-form answers. For each of the three models, CoSIR improves all seven task scores over the starting checkpoint and achieves the highest sum of scores among the compared methods. We also find that behavioral analyses demonstrate increased sharing of private information and better utilization of teammates' input, with general performance remaining close to initial checkpoints.
Our contributions are threefold:
\begin{itemize}[nosep,leftmargin=*]
    \item \textbf{Scalable collaboration training.}
    We construct a scalable training environment that preserves core collaboration behaviors with computable rewards.
    \item \textbf{Verifiable information rewards.}
    We develop team-surprisal rewards for sharing and requesting information, alongside rewards for using it.
    \item \textbf{Cross-task generalization.}
    We demonstrate transfer to seven unseen tasks across three LLM backbones while preserving general capabilities.
\end{itemize}

% \begin{figure*}[t]
% \centering
% \includegraphics[width=\textwidth]{figures/collaboration_main.png}
% \caption{Overview of collaboration training. Agents exchange privately held ratings and answer independently. An external Bayesian evaluator computes team surprisal from their evidence records, scoring statements and potential answers to requests for self-play updates. Qwen3-8B illustrates one of the three training backbones; the bars illustrate a possible posterior update. Evaluation covers unseen collaboration tasks and broader model capabilities.}
% \label{fig:main}
% \end{figure*}

\section{Related Work}
\label{sec:related-work}

\paragraph{LLM Multi-Agent Collaboration.}
Existing work on LLM multi-agent collaboration largely focuses on how agents interact at inference time. Theory-of-Mind methods help agents reason about their teammates' beliefs \citep{li2023theory}, while TalkHier organizes communication and answer refinement through a hierarchical workflow \citep{wang2025talk}. These methods guide interactions among existing models. CoSIR trains the model itself, using rewards for sharing useful information, requesting missing ratings, and using received evidence. Our experiments compare this training approach with ToM and TalkHier under matched agent counts and public speaking-turn limits.

\paragraph{Post-Training for LLM Collaboration.}
Recent studies train LLMs for multi-agent interaction. \citet{peng2025communication} use supervised fine-tuning on planning trajectories to train information sharing and seeking, together with an environment-based verifier that guides actions at inference time. MAPoRL rewards correct answers and discussion \citep{park2025maporl}. MAGRPO formulates collaboration as cooperative multi-agent reinforcement learning \citep{liu2026magrpo}, while actor-critic methods train agents for decentralized execution \citep{liu2026decentralized}. CoSIR makes the information contributed by a statement or a potential reply an explicit reward signal. The rating-table environment lets us calculate this signal from the information available to each teammate.

\paragraph{Transfer from Synthetic Multi-Agent Environments.}
Synthetic environments provide scalable opportunities for LLMs to learn through interaction. SPIRAL uses self-play in zero-sum games to develop reasoning abilities that transfer beyond the training games \citep{liu2025spiral}. MARS extends this approach to cooperative and competitive games, demonstrating transfer to unseen games and reasoning tasks \citep{yuan2025mars}. We build on the idea of learning transferable capabilities from simple environments, but target collaboration itself rather than general reasoning or game-playing strategies. In our training task, the facts agents share or request correspond to ratings in the table. This allows us to calculate how receiving a rating changes the probability of the correct option and use the result to train collaboration.
% We then test whether these behaviors generalize to unseen natural-language tasks that require agents to combine information in different ways, including reasoning across facts held by different teammates.

\section{Method}
\label{sec:method}

We design CoSIR to train LLM agents to share, request, and use information through team discussions. Our aim is for these behaviors to transfer to unseen natural-language collaboration tasks. We build a simple rating-table environment that generates tasks, gives each agent part of the information, and manages the discussion. Each task asks a team to choose the best option using ratings in a table. The environment scores the agents' messages and private answers. We use these scores to update the model, then collect new discussions with the updated model.

\subsection{A Task with Distributed Information}
\label{sec:environment}

% For each task, the environment generates a rating table for $k$ options, each described by $m$ attributes, and asks a team of $n$ agents to choose an option. The rating table $V\in\{-1,0,+1\}^{k\times m}$ contains the true ratings for this task. Entry $V_{oa}$ rates option $o$ on attribute $a$, and is uniformly distributed on $\{-1,0,+1\}$, where $-1$, $0$, and $+1$ mean poor, average, and good. For different $o$ and $a$, $V_{o,a}$'s are independent. A correct option $y^\star$ has the highest total rating:
% \begin{equation}
%     y^\star\in\arg\max_{o\in\{1,\ldots,k\}}\sum_{a=1}^{m}V_{oa}.
%     \label{eq:correct-option}
% \end{equation}
% We note that the above argmax may not be unique. We generate rating tables independently. The rating table $V$ stays fixed throughout the discussion.

For each task, the environment generates a rating table for $k$ options, each described by $m$ attributes, and asks a team of $n$ agents to choose an option. The complete table $V\in\{-1,0,+1\}^{k\times m}$ contains the ratings, where entry $V_{oa}$ rates option $o$ on attribute $a$. The values $-1$, $0$, and $+1$ mean poor, average, and good, respectively. Before filtering, the entries are sampled independently and uniformly from $\{-1,0,+1\}$. A best option $y^\star$ satisfies
\begin{equation}
    y^\star \in \arg\max_{o\in\{1,\ldots,k\}}\sum_{a=1}^{m}V_{oa}.
    \label{eq:correct-option}
\end{equation}
The maximizer need not be unique in general. In our generated tasks, we discard tables with tied highest totals. The retained table $V$ remains fixed throughout the discussion.

% Each agent receives an initial briefing containing some entries from $V$. Some ratings appear in multiple agents' briefings, while others are given to only one agent. An agent may therefore need a teammate's ratings to compare the options.
Each agent $i$ receives an initial briefing containing a subset of entries from $V$. Let $C_i\subseteq\{1,\ldots,k\}\times\{1,\ldots,m\}$ denote the observed cell indices, and define $B_i=\{((o,a),V_{oa}):(o,a)\in C_i\}$ as the observed cell--rating pairs. The sets $C_i$ may overlap, so some ratings are available to multiple agents while others are known to only one.
For example, Agent 1 may know Option A's safety rating while Agent 2 has not seen it. Agent 1 can share this rating, or Agent 2 can ask for it.

We generate tasks under five settings, summarized in \Cref{tab:configurations}. These settings specify the number of agents, which ratings each agent receives, how many ratings are given to everyone, whether agents are told which teammate holds a missing rating, and the number of discussion rounds. Each setting generates many tasks, each with its own rating table and briefings.
% \zou{what is "briefings" has not been defined}

Agents take turns sending messages in natural language. A message is the text an agent sends to its teammates on one turn. It may contain several rating statements and a request for a missing rating. Each rating statement reports a rating for one option and attribute. When received by a teammate, it is recorded as a report. During and after the discussion, we collect private answers containing an option choice and totals computed from each agent's available ratings. This allows us to check whether an agent uses the information it has received.

\paragraph{Extracting ratings and requests.}
For all three training models (Qwen3-4B, Qwen3-8B and Gemma-3-4B-it)
, we parse responses using rules and the initial Qwen3-8B model, whose parameters remain fixed. The parser extracts rating statements (option, attribute, value, and stated source), reported totals, and requests (the requested entry and any named teammate). The environment uses these results to check statements and compute rewards. \Cref{app:environment,app:parser} describe the task and parsing details.

\subsection{Measuring the Value of Information}
\label{sec:team-surprisal}

We evaluate information exchange by how it changes the probability that an option is best. Let $R_i$ denote the rating statements received by agent $i$, and define its information record as $K_i=(B_i,R_i)$. To calculate these probabilities, we represent the information in $K_i$ as a partial rating table $E_i(K_i)$, containing one rating for each available cell while leaving the remaining cells unknown. Ratings in $B_i$ take precedence; for other entries, we use the most recently received rating statement.

Let $\Omega$ be the set of complete rating tables with a unique best option, and let $\Omega_y=\{v\in\Omega:y^\star(v)=y\}$. The Bayesian evaluator uses a uniform prior $p_0(v)=1/|\Omega|$ over $v\in\Omega$. Given $K_i$, it computes the posterior probability that option $y$ is best:
\begin{equation}
    P(y\mid K_i)=\frac{\displaystyle\sum_{v\in\Omega_y}p_0(v)\prod_{(c,r)\in E_i(K_i)}\mathbf{1}[v_c=r]}{\displaystyle\sum_{v\in\Omega}p_0(v)\prod_{(c,r)\in E_i(K_i)}\mathbf{1}[v_c=r]}.
    \label{eq:briefing-probability}
\end{equation}
Here, $v$ denotes a possible complete rating table, and the indicator functions retain tables consistent with the agent's available ratings. Received rating statements are used according to their reported values; their correctness is checked separately when computing rewards. The hidden entries of the true table $V$ are not used to calculate the posterior. We compute these probabilities exactly using dynamic programming; \Cref{app:beliefs} provides the details.

For each agent, surprisal measures how unlikely the correct option is given its available information. Let $K=(K_1,\ldots,K_n)$ denote the agents' information records. We define \textbf{team surprisal} as
\begin{equation}
    \Phi(K)=-\sum_{i=1}^{n}\log_2\bar P(y^\star\mid K_i).
    \label{eq:team-surprisal}
\end{equation}
Here, $\bar P(y\mid K_i)=\max\{P(y\mid K_i),10^{-4}\}$ avoids taking the logarithm of zero.

For a set $x$ of rating statements, we measure its information value by the reduction in team surprisal if the other agents receive it:
\begin{equation}
    \Delta\Phi(x)=\Phi(K_{\mathrm{before}})-\Phi(K_{\mathrm{after}}).
    \label{eq:information-value}
\end{equation}
A positive value means the information helps the team assign a higher probability to the correct option. The value is measured in bits.

\subsection{Rewards for Communication and Answers}
\label{sec:feedback}

We use reductions in team surprisal to evaluate information sharing and requesting, and separately reward agents for using received information in their private answers. The information values defined below form the basis of our communication rewards, while Appendix~\ref{app:pricing} gives their complete computation, including reward allocation and adjustments.

\paragraph{Sharing information.}
We reward agents for sharing ratings that help their teammates identify the correct option. For a message containing rating statements $M$, we define its information value as $I_{\mathrm{share}}(M)=\Delta\Phi(M)$. This value depends on what the listeners already know. For example, suppose every agent already knows that Option B's safety rating is $+1$. Repeating this rating does not change any listener's probability of the correct option, so it provides no information reward.

Now suppose only Agent 1 knows that Option A's safety rating is $+1$. If receiving this rating increases the probability of the correct option from $1/4$ to $1/2$ for both Agent 2 and Agent 3, each listener's surprisal decreases by one bit. The total reduction is two bits before accounting for message delivery. A listener who already knows the rating gains no additional information.

A message may contain multiple rating statements whose information values are not additive. We therefore evaluate the message as a whole and allocate its information value across individual statements. Appendix~\ref{app:pricing} provides the allocation details.

However, reducing team surprisal alone is not sufficient to earn a positive reward. An agent could invent a rating that happens to increase the probability of the correct option. We therefore require a statement to be both correct and supported by the speaker's initial briefing or a received report. A correct guess without such evidence earns no positive information reward, and faithfully repeating an incorrect report also earns no positive information reward. Fabricated statements receive an additional penalty.

Conversely, even a correct rating may temporarily increase team surprisal. For example, the best option may have a poor rating on one attribute but strong ratings on others. Sharing the poor rating first could lower its probability while the other ratings remain unknown. We therefore avoid negative information rewards for statements supported by the speaker's available information. Each rating statement also incurs a small communication cost to discourage unnecessary repetition.

\paragraph{Requesting information.}
Unlike sharing a rating, asking a question does not immediately provide new information. We therefore evaluate a request by assuming that the requested rating is truthfully provided and measuring the resulting reduction in team surprisal, using the same measure as for sharing. Let $q$ be a teammate holding the requested cell $c$, and let $x_{q,c}$ denote a hypothetical reply containing its true rating. We define the potential information value of the request as $I_{\mathrm{request}}(q,c)=[\Delta\Phi(x_{q,c})]_+$, where $[z]_+=\max(z,0)$.

Specifically, the request must concern a rating missing from the requester's initial briefing but known to a teammate. The environment uses the true value from $V$ to construct the hypothetical reply. We then calculate how receiving it would change the probability of the correct option for every agent other than the replying teammate, including the requester. The potential value is calculated when the question is asked, without waiting for an actual reply.

For example, before Agent 1 shares Option A's safety rating, Agent 2 could ask for it. If the truthful reply of $+1$ increases the probability of the correct option from $1/4$ to $1/2$ for both Agent 2 and Agent 3, the request has a potential information value of two bits. The final reward also accounts for previously received information, repeated requests, and whether the holder has another opportunity to speak. Appendix~\ref{app:pricing} provides these adjustments.

\paragraph{Using received information.}
Sharing and requesting information are useful only if agents can also use what they receive. We therefore evaluate two parts of each private answer: the selected option and the calculated total ratings for all options.

Let $\hat y_i$ be agent $i$'s selected option, $\hat s_{io}$ its reported total for option $o$, and $s^K_{io}$ the reference total computed from its available ratings. Each table entry is counted once, and unknown ratings are treated as zero. For a parseable choice, the private-answer reward is
\begin{equation}
    r_i^{\mathrm{answer}} = \bar P(\hat y_i\mid K_i) + \frac{1}{k}\sum_{o=1}^{k}\mathbf{1}[\hat s_{io}=s^K_{io}].
    \label{eq:answer-reward}
\end{equation}
The choice score uses the probability that the selected option is best given the agent's available information, rather than directly comparing it with the true answer. The calculation score is the fraction of options whose reported totals match the reference totals. Missing or unreadable totals count as incorrect.

For example, suppose an agent initially does not know Option A's safety rating and then receives a report that it is $+1$. The evaluator uses this report to update the option probabilities, while the reference total for Option A increases by one. We evaluate the agent's next private answer using both the updated probabilities and totals. The two scores thus encourage agents to use received information both when choosing an option and when calculating its total rating. Appendix~\ref{app:answer-targets} provides the complete scoring rules, including how conflicting reports are handled.

\subsection{Learning from Team Discussions}
\label{sec:training}

We use the sharing, requesting, and private-answer rewards to train LLMs through self-play, where the same model plays all agents in a team. At each speaking turn, we sample multiple candidate responses from the same briefing and conversation, allowing us to compare different responses under the same conditions. We score all candidates and randomly select one to continue the discussion, while the others also contribute to training. Private answers are sampled and scored separately. After collecting discussions, we update the model and repeat the process with the updated policy.

A message may contain useful ratings, repetitions, and questions, so we assign rewards to the corresponding parts of the response rather than to the entire message. Statement rewards provide direct credit to their corresponding tokens, while request and private-answer scores are converted into group-relative advantages. These signals guide updates to corresponding tokens through a variant of PPO \citep{schulman2017ppo}.

Early in training, the model may rarely generate useful questions. When the evaluator identifies a request as the best available action, we add a few demonstrations expressing that request. These examples are scored alongside sampled responses and trained with a clipped imitation update, but never selected to continue the discussion.

To help preserve general capabilities, we use a reference-model penalty and train on responses generated by the initial checkpoint for general prompts. Appendix~\ref{app:training-details} provides the sampling, token-level credit assignment, demonstration generation, and optimization details.

\section{Experiments}
\label{sec:experiments}

% Styling shared by the experimental tables.
\definecolor{resultblue}{RGB}{31,78,121}
\definecolor{resultshade}{RGB}{239,244,248}
\newcommand{\resultband}{\rlap{\textcolor{resultshade}{\rule[-\arraystretch\dp\strutbox]{\linewidth}{\dimexpr\arraystretch\ht\strutbox+\arraystretch\dp\strutbox\relax}}}}
\newcommand{\resultpair}[2]{\makebox[2.1em][r]{#1}\,\ensuremath{\rightarrow}\,\textcolor{resultblue}{\makebox[2.1em][r]{#2}}}

We evaluate transfer to unseen collaboration tasks, changes in agents' communication, and retention of general capabilities.

\subsection{Experimental Setup}
\label{sec:experimental-setup}

\paragraph{Models and baselines.}
We evaluate Qwen3-8B and Qwen3-4B \citep{yang2025qwen3}, and Gemma-3-4B-it \citep{gemmateam2025gemma3}. Base is the starting checkpoint, and CoSIR is the checkpoint after collaboration training. Both use the same task prompts and evaluation settings. Outcome-RL uses the same self-play environment and model initialization as CoSIR, but learns from the final team outcome rather than behavior-specific information rewards. It retains the request demonstrations, while outcome feedback is assigned to the sampled discussion trajectory. This comparison tests the reward design within outcome-based multi-agent reinforcement learning \citep{liu2026magrpo}. We also evaluate Theory of Mind (ToM) \citep{li2023theory} and TalkHier \citep{wang2025talk}, which add reasoning or revision steps at inference time. They use the same numbers of agents and speaking turns, with additional model calls described in \Cref{app:evaluation-details}.

\paragraph{Unseen collaboration tasks.}
We evaluate on seven tasks outside the training environment. HiddenBench \citep{li2026hiddenbench} requires decisions from private facts, and SILO-BENCH \citep{zhang2026silo} requires computation over data divided among agents. SplitQA asks agents to connect facts from separate documents. MathChat \citep{wu2023mathchat} evaluates interactive mathematical problem solving, while BenchForm \citep{weng2025conformity} evaluates answers under peer influence. Einstein Puzzles \citep{peng2025communication} and Collaboration Tax \citep{sun2026collaboration} evaluate further forms of reasoning and coordination involving distributed information or task responsibilities. \Cref{app:evaluation-details} gives the protocols, generation settings, and scoring rules.

We construct SplitQA to test whether agents can use distributed evidence to generate an answer. In HiddenBench, private facts support or rule out the listed options. SplitQA instead uses HotpotQA questions \citep{yang2018hotpotqa} whose answers require information from two supporting paragraphs. We assign those paragraphs to different agents and give distractors to the other two. Agents follow the HiddenBench discussion protocol and then provide free-form answers. \Cref{app:splitqa} describes the 70-question test set and its scoring.

\paragraph{Reporting.}
We report mean scores and standard deviations over four runs per task. SUM adds the seven task means with equal weight, regardless of sample count. \Cref{app:evaluation-details} provides the scoring and aggregation details. We separately evaluate mathematical reasoning on GSM8K \citep{cobbe2021verifiers}, general knowledge on MMLU \citep{hendrycks2021mmlu}, and instruction following on IFEval \citep{zhou2023ifeval}.

\begin{table*}[t]
\centering\small
\setlength{\tabcolsep}{1.8pt}
\renewcommand{\arraystretch}{1.12}
\begin{tabular*}{\textwidth}{@{\extracolsep{\fill}}lrrrrrrrrr@{}}
\toprule
& \multicolumn{7}{c}{Unseen collaboration tasks: mean $\pm$ std} & \multicolumn{2}{c}{Overall transfer} \\
\cmidrule(lr){2-8}\cmidrule(l){9-10}
Method & \shortstack{Hidden\\Bench} & SILO & SplitQA & MathChat & \shortstack{Bench\\Form} & Einstein & \shortstack{Collab.\\Tax} & \textbf{SUM} & $\Delta$SUM \\
\midrule
\multicolumn{10}{@{}l}{\textbf{Qwen3-8B}} \\
Base & \meansd{64.23}{3.85} & \meansd{63.53}{3.56} & \meansd{57.50}{6.32} & \meansd{78.48}{1.97} & \meansd{49.80}{0.99} & \meansd{3.58}{0.69} & \meansd{35.48}{7.44} & 352.60 & 0.00 \\
ToM & \meansd{68.85}{4.24} & \meansd{\mathbf{71.70}}{3.99} & \meansd{56.43}{1.84} & \meansd{73.95}{0.20} & \meansd{51.05}{3.89} & \meansd{5.42}{0.17} & \meansd{26.78}{2.92} & 354.17 & +1.57 \\
TalkHier & \meansd{65.38}{5.25} & \meansd{68.63}{1.60} & \meansd{53.93}{0.71} & \meansd{81.96}{2.70} & \meansd{\mathbf{56.80}}{5.37} & \meansd{2.50}{0.33} & \meansd{35.45}{1.14} & 364.65 & +12.06 \\
Outcome-RL & \meansd{67.31}{4.76} & \meansd{68.25}{3.54} & \meansd{57.50}{2.95} & \meansd{79.55}{3.61} & \meansd{51.80}{1.98} & \meansd{4.33}{1.09} & \meansd{34.98}{3.92} & 363.72 & +11.12 \\
\resultband\textcolor{resultblue}{\textbf{CoSIR}} & \meansd{\mathbf{75.00}}{2.31} & \meansd{68.93}{2.36} & \meansd{\mathbf{60.00}}{4.21} & \meansd{\mathbf{82.45}}{2.24} & \meansd{53.05}{0.35} & \meansd{\mathbf{6.25}}{1.26} & \meansd{\mathbf{38.78}}{4.39} & \textcolor{resultblue}{\textbf{384.44}} & \textcolor{resultblue}{+31.85} \\
\midrule
\multicolumn{10}{@{}l}{\textbf{Qwen3-4B}} \\
Base & \meansd{64.62}{3.32} & \meansd{67.00}{5.23} & \meansd{57.50}{3.76} & \meansd{75.04}{1.52} & \meansd{46.20}{0.57} & \meansd{5.42}{0.17} & \meansd{17.40}{3.81} & 333.17 & 0.00 \\
ToM & \meansd{65.00}{5.67} & \meansd{67.50}{4.43} & \meansd{49.29}{3.40} & \meansd{68.96}{0.00} & \meansd{47.40}{1.13} & \meansd{1.25}{0.83} & \meansd{19.83}{2.19} & 319.22 & -13.95 \\
TalkHier & \meansd{58.08}{9.60} & \meansd{61.75}{5.06} & \meansd{\mathbf{60.36}}{5.52} & \meansd{68.96}{1.33} & \meansd{\mathbf{48.45}}{2.62} & \meansd{2.83}{0.43} & \meansd{21.33}{3.75} & 321.75 & -11.42 \\
Outcome-RL & \meansd{66.54}{5.94} & \meansd{68.50}{2.38} & \meansd{57.14}{4.21} & \meansd{\mathbf{76.55}}{2.21} & \meansd{41.85}{2.33} & \meansd{4.33}{0.00} & \meansd{20.00}{1.33} & 334.92 & +1.74 \\
\resultband\textcolor{resultblue}{\textbf{CoSIR}} & \meansd{\mathbf{71.92}}{4.42} & \meansd{\mathbf{73.75}}{2.99} & \meansd{57.86}{5.02} & \meansd{\mathbf{76.55}}{2.06} & \meansd{47.65}{0.78} & \meansd{\mathbf{7.92}}{0.17} & \meansd{\mathbf{22.23}}{1.16} & \textcolor{resultblue}{\textbf{357.87}} & \textcolor{resultblue}{+24.70} \\
\midrule
\multicolumn{10}{@{}l}{\textbf{Gemma-3-4B-it}} \\
Base & \meansd{33.85}{5.62} & \meansd{33.25}{1.26} & \meansd{20.00}{2.61} & \meansd{70.47}{0.89} & \meansd{45.10}{0.42} & \meansd{0.00}{0.00} & \meansd{5.35}{1.68} & 208.02 & 0.00 \\
ToM & \meansd{32.69}{5.08} & \meansd{22.50}{4.12} & \meansd{22.50}{2.14} & \meansd{57.97}{2.80} & \meansd{45.10}{0.00} & \meansd{\mathbf{0.33}}{0.00} & \meansd{7.38}{2.83} & 188.47 & -19.55 \\
TalkHier & \meansd{35.77}{6.57} & \meansd{25.00}{2.16} & \meansd{22.50}{6.32} & \meansd{64.53}{0.86} & \meansd{\mathbf{47.20}}{0.00} & \meansd{\mathbf{0.33}}{0.38} & \meansd{\mathbf{8.18}}{5.05} & 203.51 & -4.50 \\
Outcome-RL & \meansd{\mathbf{41.15}}{2.63} & \meansd{34.00}{2.71} & \meansd{17.86}{6.00} & \meansd{66.97}{1.45} & \meansd{45.10}{0.42} & \meansd{0.00}{0.00} & \meansd{6.00}{5.28} & 211.08 & +3.06 \\
\resultband\textcolor{resultblue}{\textbf{CoSIR}} & \meansd{40.00}{2.18} & \meansd{\mathbf{39.00}}{3.65} & \meansd{\mathbf{28.21}}{2.95} & \meansd{\mathbf{74.54}}{0.73} & \meansd{45.70}{0.14} & \meansd{\mathbf{0.33}}{0.38} & \meansd{8.15}{3.36} & \textcolor{resultblue}{\textbf{235.94}} & \textcolor{resultblue}{+27.92} \\
\bottomrule
\end{tabular*}
\caption{Transfer to seven unseen collaboration tasks. Task scores (\%) show mean $\pm$ standard deviation. SUM is the sum of seven task scores, each measured on a 0–100 scale. $\Delta$SUM is the change from Base for the same model. Shading marks CoSIR. Bold marks the best score within each model, including ties. SILO denotes SILO-BENCH and Collab.\ Tax denotes Collaboration Tax.}
\label{tab:ood-results}
\end{table*}

\subsection{Generalization Across Tasks and Models}
\label{sec:ood-results}

\Cref{tab:ood-results} shows that CoSIR achieves the highest SUM for all three models, exceeding the strongest alternative by 19.79, 22.95, and 24.86 points for Qwen3-8B, Qwen3-4B, and Gemma-3-4B-it, respectively. Each trained model also achieves a higher mean score than Base on all seven tasks, indicating that the gains are not driven by a single benchmark.

Inference-time methods perform better on some individual tasks. ToM leads on SILO-BENCH with Qwen3-8B, while TalkHier leads on BenchForm with all three models. However, both methods reduce SUM relative to Base for Qwen3-4B and Gemma-3-4B-it, whereas CoSIR improves SUM across all three models.

SplitQA further tests transfer beyond choosing among predefined options, requiring agents to combine facts from different supporting paragraphs to produce a free-form answer. CoSIR improves over Base by 2.50, 0.36, and 8.21 points for Qwen3-8B, Qwen3-4B, and Gemma-3-4B-it, respectively. The improvement is small for Qwen3-4B, but the results suggest that the learned behaviors can also help agents combine evidence across documents.

\begin{table*}[t]
\centering\small
\setlength{\tabcolsep}{4pt}
\renewcommand{\arraystretch}{1.07}
\begin{tabular*}{\textwidth}{@{\extracolsep{\fill}}llrrr@{}}
\toprule
Task & Behavior & Qwen3-8B & Qwen3-4B & Gemma-3-4B-it \\
\midrule
HiddenBench & Private facts disclosed $\uparrow$ & \resultpair{68.5}{74.3} & \resultpair{24.5}{33.9} & \resultpair{9.9}{15.3} \\
& Messages responding to teammates $\uparrow$ & \resultpair{59.4}{65.8} & \resultpair{64.0}{73.4} & \resultpair{93.1}{95.4} \\
\addlinespace[2pt]
SILO-BENCH & Agents listing all own data $\uparrow$ & \resultpair{55.3}{65.4} & \resultpair{74.8}{78.4} & \resultpair{76.8}{85.9} \\
& Result-only messages $\downarrow$ & \resultpair{16.2}{12.8} & \resultpair{10.2}{8.1} & \resultpair{10.6}{9.3} \\
\addlinespace[2pt]
SplitQA & Holders stating supporting facts $\uparrow$ & \resultpair{52.9}{58.2} & \resultpair{56.8}{62.9} & \resultpair{33.6}{35.7} \\
\addlinespace[2pt]
MathChat & Messages with correction language $\uparrow$ & \resultpair{2.5}{11.1} & \resultpair{4.9}{6.2} & \resultpair{40.1}{44.4} \\
\bottomrule
\end{tabular*}
\caption{Observed collaboration behaviors (\%). Each pair reads Base $\rightarrow$ \textcolor{resultblue}{CoSIR}. Arrows beside the measures indicate the intended direction. Disclosure measures cover any discussion round. Correction language records attempts to correct an error, without judging whether the correction succeeds. \Cref{app:behavior-details} defines the denominators.}
\label{tab:behavior-results}
\end{table*}

\paragraph{Comparison with Outcome-RL.}
Outcome-RL follows the same training procedure as CoSIR but rewards only the final team outcome. Although it improves SUM over Base for all three models, CoSIR exceeds it by 20.72, 22.95, and 24.86 points for Qwen3-8B, Qwen3-4B, and Gemma-3-4B-it, respectively. CoSIR scores higher on all seven tasks with Qwen3-8B and six tasks with each of the other models. It ties Outcome-RL on MathChat with Qwen3-4B, while Outcome-RL performs better on HiddenBench with Gemma-3-4B-it.

Unlike Outcome-RL, which evaluates only the final team result, CoSIR rewards individual rating statements and requests, as well as choices and calculations in private answers. Thus, responses leading to the same final outcome can receive different feedback based on how they share and use information. These results support the benefit of fine-grained information rewards over final-outcome feedback within our self-play training framework. We next examine how agents' communication changes after training.

\begin{table}[t]
\centering\small
\setlength{\tabcolsep}{3pt}
\renewcommand{\arraystretch}{1.04}
\begin{tabular*}{\columnwidth}{@{\extracolsep{\fill}}lrr@{}}
\toprule
Failure type & Base & \textcolor{resultblue}{CoSIR} \\
\midrule
\resultband\textit{Specification and system design} & \textbf{88.2} & 88.3 \\
\quad Disobey task specification & \textbf{5.6} & 6.1 \\
\quad Disobey role specification & 0.0 & 0.0 \\
\quad Repeated steps & \textbf{85.1} & 85.6 \\
\quad Loss of conversation history & \textbf{9.5} & 9.8 \\
\quad Unaware of when to stop & \textbf{18.5} & 22.4 \\
\addlinespace[3pt]
\resultband\textit{Coordination between agents} & 67.8 & \textbf{62.0} \\
\quad Conversation reset & \textbf{0.0} & 0.1 \\
\quad Proceed without clarification & 36.4 & \textbf{32.0} \\
\quad Task derailment & \textbf{7.9} & 8.5 \\
\quad Information withholding & \textbf{18.5} & 18.7 \\
\quad Ignore other agents' input & 26.3 & \textbf{22.8} \\
\quad Reasoning and action mismatch & 42.5 & \textbf{38.8} \\
\addlinespace[3pt]
\resultband\textit{Verification and termination} & 58.8 & \textbf{53.5} \\
\quad Premature termination & \textbf{6.8} & 8.8 \\
\quad No or incomplete verification & 51.3 & \textbf{45.9} \\
\quad Incorrect verification & \textbf{20.1} & 20.8 \\
\midrule
Failure types per conversation & 3.29 & \textbf{3.20} \\
\bottomrule
\end{tabular*}
\caption{MAST failures (\%) pooled across three models on HiddenBench, SplitQA, and MathChat \citep{cemri2025fail}. Shaded rows give category totals, and bold marks the lower value. The final row gives the mean count of failure types. \Cref{app:mast-results} defines the aggregation.}
\label{tab:mast-results}
\end{table}

\subsection{How Collaboration Changes}
\label{sec:behavior-analysis}

Higher task scores do not describe how a discussion improves. We therefore examine whether agents make their private information available and respond to their teammates. \Cref{tab:behavior-results} reports these behaviors before and after training. \Cref{app:behavior-details} defines the measures.

\paragraph{Making evidence available.}
After training, agents share more private information across all three models on HiddenBench. For Qwen3-8B, the disclosure rate rises from 68.5\% to 74.3\%, and the fraction of messages responding to teammates increases from 59.4\% to 65.8\%. On SILO-BENCH, more agents share their complete local data, while fewer messages report only an accumulated result, making the underlying data available for verification. On SplitQA, agents share their supporting facts more often, while MathChat shows more messages attempting to identify or correct errors.

Agents also share information earlier. For Qwen3-8B, first-message disclosure increases from 61.1\% to 69.0\% on HiddenBench and from 41.1\% to 51.1\% on SplitQA (\Cref{app:behavior-details}). Unlike the any-round measures in \Cref{tab:behavior-results}, these figures capture information shared early enough for teammates to use in later turns.

\paragraph{Using teammates' information.}
Sharing information helps only if teammates use it. We therefore apply the Multi-Agent System Failure Taxonomy (MAST) of \citet{cemri2025fail} to conversations from HiddenBench, SplitQA, and MathChat. \Cref{tab:mast-results} reports failure rates pooled across the three models. Coordination failures decrease from 67.8\% to 62.0\%, and verification or termination failures from 58.8\% to 53.5\%. In contrast, specification and system-design failures remain nearly unchanged (88.2\% versus 88.3\%), consistent with CoSIR's focus on sharing, requesting, and using information rather than enforcing task specifications or managing discussion procedures. Among individual failure types, ignoring teammates' input decreases from 26.3\% to 22.8\%, proceeding without needed clarification from 36.4\% to 32.0\%, and missing or incomplete verification from 51.3\% to 45.9\%. These changes complement the disclosure measures by showing fewer failures to use teammates' information or check the team's answers.

However, repeated steps (85.1\% versus 85.6\%) and information withholding (18.5\% versus 18.7\%) do not improve. These results suggest that CoSIR improves some aspects of collaboration without addressing every failure type.

\paragraph{From disclosure to a better answer.}
Three paired Qwen3-8B cases in \Cref{app:case-studies} illustrate how information changes decisions: HiddenBench shows a revised recommendation, SplitQA combines distributed facts, and SILO-BENCH shows a corrected calculation using shared data. 
% The team must count occurrences of \texttt{delta} in 16 words divided among four agents. Base agents repeat their local counts without providing the word lists, and all finish with the incorrect total of 18. After training, Agent 3 reports a count of zero but also shares \texttt{delta, delta, beta, epsilon}. Agent 1 checks the list, corrects that count to two, and obtains $2+1+2+0=5$. All four agents then give the correct total. Sharing the words gives the teammate enough information to check the reported count and correct the total. 
% \Cref{app:case-studies} provides the dialogue excerpts.

\paragraph{General capabilities.}
Across GSM8K, MMLU, and IFEval, performance changes from Base range from $-0.6$ to $+0.8$ percentage points, indicating that general capabilities remain largely preserved (\Cref{app:general-capabilities}).

\section{Conclusion}
\label{sec:conclusion}

CoSIR trains LLMs to share, request, and use information with computable rewards in a rating-table environment. Across three backbones, the trained models achieve higher mean scores than Base on all seven unseen collaboration tasks and the highest aggregate scores among the compared methods. Behavioral analyses and case studies show more sharing and use of teammates' information, while general capabilities remain close to the starting checkpoints.

\clearpage
\section*{Limitations}

Our evaluation covers three model backbones and seven collaboration tasks. The trained models improve over their starting checkpoints throughout this suite, but inference-time workflows and Outcome-RL remain stronger on some combinations of model and task. Training also does not remove every communication failure: repeated steps and several other MAST failure categories remain frequent. The reward depends on a toy environment with known facts and a computable evaluator. Extending the same scoring approach to open-ended tasks would require a way to assess information value when the complete state is unavailable.

\bibliography{custom}

\clearpage
\appendix
\crefalias{section}{appendix}

\section{Environment Construction}
\label{app:environment}

\subsection{Ratings and briefings}

Every instance has four options and five attributes, with two, three, or four agents. Option names, attribute names, and decision scenarios vary across instances, while the highest-total rule remains fixed. Cell ratings are initially sampled independently and uniformly from $\{-1,0,+1\}$, and tables with tied highest totals are rejected. The generator then applies the conditions for the selected information configuration. These conditions shape which ratings agents need to exchange. The Bayesian evaluator does not use these conditions as assumptions.

Let $\mathcal C=\{1,\ldots,k\}\times\{1,\ldots,m\}$ be the table cells and $C_i\subseteq\mathcal C$ the cells in agent $i$'s briefing. Define
\begin{equation}
\begin{split}
    B_i&=\{(c,V_c):c\in C_i\},\\
    S&=\bigcap_{i=1}^n C_i,\qquad
    U_i=C_i\setminus\bigcup_{j\ne i}C_j.
\end{split}
\label{eq:information-allocation}
\end{equation}
The set $S$ contains ratings initially shared by every agent. A rating in $U_i$ appears only in agent $i$'s briefing. Cells outside $C_{\mathrm{team}}=\bigcup_i C_i$ are initially unknown to everyone.

\begin{table*}[t]
\centering
\small
\begin{tabular}{@{}cccp{0.40\textwidth}ccc@{}}
\toprule
Family & Weight & Agents & Private ratings & Shared & Holders shown & Rounds \\
\midrule
A & 30\% & 4 & Usually one cell per agent & 3--5 & No & 3--6 \\
B & 15\% & 2 & Usually two cells per agent & 3--5 & No & 3--5 \\
C & 20\% & 4 & Portions of sizes $(8,4,4,4)$ & 0 & No & 3--6 \\
D & 15\% & 2 & $(6,6)$, $(7,5)$, or $(8,4)$ cells & 4 & Yes & 4--8 \\
E & 20\% & 3 & $(5,5,5)$, $(7,5,3)$, or $(6,6,2)$ cells & 4 & No & 4--8 \\
\bottomrule
\end{tabular}
\caption{Information configurations. Parenthesized counts give the ratings allocated across agents. The number of discussion rounds is sampled uniformly within each listed range.}
\label{tab:configurations}
\end{table*}

Families A and B use shared ratings that favor an incorrect option. Their checks rank options by the ratings in a specified cell set, treating missing ratings as zero. For $C\subseteq\mathcal C$, write
\begin{equation}
    \begin{aligned}
    s_o(C)&=\sum_{a:(o,a)\in C}V_{oa},\\
    \operatorname{best}(C)&=\arg\max_o s_o(C).
    \end{aligned}
    \label{eq:known-score-order}
\end{equation}
Here $\operatorname{best}(C)$ is a set, so it includes all ties. With $y^{\mathrm{decoy}}$ denoting the incorrect option favored by shared ratings, these two families require
\begin{equation}
\begin{aligned}
    \operatorname{best}(C_{\mathrm{team}})&=\{y^\star\},\\
    \operatorname{best}(S)&=\{y^{\mathrm{decoy}}\},\quad y^{\mathrm{decoy}}\ne y^\star,\\
    \operatorname{best}(C_i)&\ne\{y^\star\}\quad\text{for all }i,\\
    \operatorname{best}(C_{\mathrm{team}}\setminus\{c\})&\ne\{y^\star\},\\
    &\hspace{-12mm}\text{for all }c\in\textstyle\bigcup_i U_i.
\end{aligned}
\label{eq:briefing-constraints}
\end{equation}
Thus, pooling the assigned ratings gives the correct option a unique lead, while removing any exclusive cell removes that lead. These checks use known-rating totals, not the option with the highest evaluator probability for each agent.

Family C has no shared cells and applies its checks to each agent's portion of the table. The evaluator's highest-probability option for an individual portion must differ from the correct option. Removing any one portion must also leave the correct option without a unique probability lead under the remaining information. This checks whole portions rather than requiring each private cell to be necessary.

Families D and E provide longer briefings and allow an individual agent's highest-probability option to be correct. About 70\% of their instances additionally check the need to combine agents' private ratings, removing one agent's private portion at a time. The remaining 30\% use random tables without that extra condition. Even pooled known-rating totals need not always select the correct option uniquely. The full table still has a unique best option. None of these family conditions enters the evaluator's prior.

\section{Communication Protocol and Parsing}
\label{app:parser}

\subsection{Discussion and request routing}
\label{app:conversation}

A message is one complete natural-language utterance. It may contain several statements, each claiming a rating for one cell, and at most one request for a rating. Requests can name a teammate or address the group.

Let $T$ be the total number of discussion rounds. It is drawn uniformly from the selected family's interval in \Cref{tab:configurations}. Agents see the current round and whose turn it is, but not $T$ or the number of remaining rounds. The speaking order is fixed.

A request identifies one cell and may name a recipient. Outside family D, a group-addressed request is routed to the next holder in speaking order. A request naming a teammate who cannot provide a useful answer can also be rerouted. Family D shows holders and retains the named target. Routing uses the environment's allocation record, but does not insert the holder's rating into the requester's prompt. It is a training-time forwarding mechanism that uses information hidden from the agents.

For each ordinary listener, the environment uses a deterministic hash of the game, turn, and listener to draw a message-delivery outcome. The miss probability is $0.3$. All 16 candidate messages at one speaking state share this ordinary delivery mask. If a candidate has a routed request, its recipient receives the full message regardless of that mask. Future prompts include only messages actually delivered to that agent. No agent is told which other listeners missed a message.

A request remains pending until its recipient reports the requested cell or the requester leaves. The recipient is not forced to reply. Leaving an answerable request pending incurs the action penalty on each relevant speaking turn. Reporting the cell clears the request even if the reported value is fabricated. The statement reward separately checks whether the value is supported. For a given requester, a new request for a cell already pending earns no information credit, even if it names a different recipient. This repetition is not itself classified as an invalid request.

Only the current speaker answers privately before its turn. After the last round, every agent answers privately. Each answer uses that agent's briefing and delivered conversation. It is neither broadcast nor added to rating statement records. Intermediate and final answers form separate candidate groups under the same scoring rules.

\subsection{Text extraction and information checks}

The parser combines frozen initial Qwen3-8B with rules, using temperature zero, disabled thinking, and prompt caching. It is shared across all three training backbones. Rules first locate option names, attribute names, and rating words in clauses or tables. The model extracts statements, claimed sources, totals, and an optional request with its target. Clear request phrases take precedence when identifying the intended action.

Text checks remove conditional or negated claims, unknown-value placeholders, and ratings that do not match the corresponding clause. A small set of fixed sentence patterns recovers missed ratings. If model extraction is unusable, rule-extracted ratings that pass the checks remain available. Empty messages yield no claims. These checks establish what the text says, not whether a rating is true.

The environment then compares each extracted statement with the speaker's briefing and actually received rating statements. A matching briefing value is an own statement. A matching received rating statement can be relayed with its original source and identifier. A rating with neither basis is unsupported, even if it equals the hidden table value. Checking truth against the table is a separate operation used in reward calculation.

\subsection{Rating statement identity and delivery}

For each rating statement, we store a cell, a rating, an original speaker, and a stable identifier. A relay copies the original rating statement rather than creating a new observation from the relaying agent. Fabricated rating statements use identifiers determined by the speaker, cell, and rating. Repeating the same fabricated statement does not create a new report or change the rating retained for posterior computation. Report identifiers prevent repeated messages from being counted as new reports or changing their original reception times. For posterior computation, the evaluator uses at most one rating per cell, following the precedence rule described in Appendix C.

Let $R_j^t$ be listener $j$'s rating statements before turn $t$, $M_t$ the rating statements in the current message, and $D_{tj}$ its delivery indicator. The update is
\begin{equation}
\begin{split}
    R_j^{t+1}&=\begin{cases}
        R_j^t\cup M_t,&D_{tj}=1,\\
        R_j^t,&D_{tj}=0,
    \end{cases}\\
    K_j^{t+1}&=(B_j,R_j^{t+1}),\qquad j\ne i,
\end{split}
\label{eq:evidence-update}
\end{equation}
for speaker $i$. The union is by identifier: a repeated identifier neither adds a rating statement nor refreshes its first-reception time. Rating statements are not added back to their original speaker. Unsupported rating statements can enter a listener's record. Delivery does not replace them with true ratings.

\section{Bayesian Evaluator}
\label{app:beliefs}

\subsection{Prior and Observations}

The evaluator considers possible tables consistent with an agent's initial briefing and assigns probabilities based on the rating statements it has received. Let $\Omega$ be the tables with a unique highest-scoring option. Independent uniform cell ratings, conditioned on this requirement, give
\begin{equation}
\begin{split}
    \Omega&=\left\{v\in\{-1,0,+1\}^{k\times m}:\right.\\
        &\hspace{10mm}\left.\left|\arg\max_o\sum_a v_{oa}\right|=1\right\},\\
    p_0(v)&=1/|\Omega|,\qquad v\in\Omega.
\end{split}
\label{eq:evaluator-prior}
\end{equation}
This simplified prior excludes the generator's family-specific rejection conditions. The posterior is exact for this evaluation model, not for the full training-instance distribution.

For each agent, the information record is $K_i=(B_i,R_i)$, where $B_i$ contains the initial observations and $R_i$ contains received rating statements. We construct $E_i(K_i)$ by retaining one rating per cell. An initial briefing value takes precedence; otherwise, we use the most recently received rating for that cell. Repeated reports with the same identifier do not change the retained value or its reception time.

The evaluator conditions on these observed ratings without assigning a fixed reliability probability to received statements. Incorrect reports may therefore affect the resulting posterior. Their correctness is checked separately against the true table $V$ when assigning statement rewards.

\subsection{Exact posterior computation}

For a possible table $v$, the unnormalized posterior weight is the prior restricted to tables consistent with the agent's available ratings:
\begin{equation}
    W_i(v)=p_0(v)\prod_{(c,r)\in E_i(K_i)}\mathbf{1}[v_c=r].
\end{equation}
Let $y^\star(v)$ be that table's unique best option. The desired posterior is
\begin{equation}
    P(y\mid K_i)=\frac{\sum_{v\in\Omega}\mathbf1[y^\star(v)=y]W_i(v)}
    {\sum_{v\in\Omega}W_i(v)}.
    \label{eq:evaluator-posterior}
\end{equation}
The actual instance's hidden label $y^\star$ is absent from this computation. It is used only afterward to score communication.

Dynamic programming avoids enumerating all tables. Before conditioning on a unique winner, form a normalized distribution $q_{ioa}(u)$ for each cell value. For each cell, $q_{ioa}(u)$ is a point mass at the retained rating if the cell appears in $E_i(K_i)$, and is uniform over $\{-1,0,+1\}$ otherwise. Let $F_{io}^{(a)}(s)$ be the probability that the first $a$ ratings of option $o$ sum to $s$. Starting from $F_{io}^{(0)}(0)=1$ and zero elsewhere, compute
\begin{equation}
    F_{io}^{(a)}(s)=\sum_{u\in\{-1,0,+1\}}q_{ioa}(u)F_{io}^{(a-1)}(s-u).
    \label{eq:row-sum-dp}
\end{equation}
Write $F_{io}=F_{io}^{(m)}$ for the full row-sum distribution and $H_{io}(s)=\sum_{z\le s}F_{io}(z)$ for its cumulative probability. Row sums range from $-m$ to $m$. The probability that option $y$ strictly exceeds every other option, before conditioning on uniqueness, is
\begin{equation}
\begin{split}
    b_{iy}&=\sum_{s=-m}^{m}F_{iy}(s)\prod_{o\ne y}H_{io}(s-1),\\
    P(y\mid K_i)&=\frac{b_{iy}}{\sum_{o=1}^k b_{io}}.
\end{split}
\label{eq:strict-winner-dp}
\end{equation}
The denominator removes tables with a tied maximum. After the cell distributions are formed, row convolution takes $O(km^2)$ operations, and direct winner calculation takes $O(k^2m)$. This is a finite exact sum, not a Monte Carlo estimate.

For numerical scoring, we use $\bar P(y\mid K_i)=\max\{P(y\mid K_i),10^{-4}\}$, as in \Cref{eq:team-surprisal}. This bounds each agent's surprisal at approximately $13.3$ bits. The floored values need not sum to one, while $P$ remains normalized. A reduction in $\Phi(K)$ need not increase every agent's probability or the LLM's actual decision accuracy. The communication rewards use these floored values.

\section{Reward Computation}
\label{app:pricing}

\subsection{Statement reward computation}

For a speaker $i$ and a set of rating statements $M$, let $K_j\oplus M$ add rating statements with previously unseen identifiers to listener $j$'s record. We compute each listener's contribution to the decrease in team surprisal (\Cref{eq:team-surprisal}) using the floored probabilities from \Cref{app:beliefs}:
\begin{equation}
\begin{aligned}
    \bar\Delta_j(M)&=\log_2\frac{\bar P(y^\star\mid K_j\oplus M)}{\bar P(y^\star\mid K_j)},\\
    \mathcal V_i(M;K)&=0.7\sum_{j\ne i}\bar\Delta_j(M).
\end{aligned}
\label{eq:message-value}
\end{equation}
The sum covers present listeners other than the speaker. Rating statements they already hold cause no change. The uniform nominal delivery weight $0.7$ matches ordinary message delivery, but also applies to a routed request recipient who is guaranteed to receive the whole message. Thus, the message value is not the exact expected change under every delivery condition. Actual rating statement records follow realized delivery. The scoring weight and the state-update rule serve different purposes.

Partition a message's statements into supported statements $L$ and unsupported statements $I$. For fixed speaker and information records, abbreviate the value of these rating statements as $\mathcal V(L)=\mathcal V_i(L;K)$. Let $a_\ell$ denote the value allocated to statement $\ell$. For $\ell\in L$, start with the removal marginal
$m_\ell=\mathcal V(L)-\mathcal V(L\setminus\{\ell\})$.
Let $Z_+=\sum_{h:m_h>0}m_h$ and $Z_-=\sum_{h:m_h\le0}m_h$. If $Z_+>0$ and $\mathcal V(L)-Z_-\ge0$, assign
\begin{equation}
    a_\ell=\begin{cases}
        m_\ell\bigl(\mathcal V(L)-Z_-\bigr)/Z_+,&m_\ell>0,\\
        m_\ell,&m_\ell\le0.
    \end{cases}
    \label{eq:statement-allocation}
\end{equation}
This keeps negative marginals unchanged and rescales positive marginals so that allocated scores sum to $\mathcal V(L)$. A single supported statement receives its own message value.

When this rescaling is not applicable, the allocation averages sequential marginal values over statement orders for sets of at most six statements. This is the Shapley allocation \citep{shapley1953value}: each statement receives its average change in value when added after the statements preceding it. For larger sets, the fallback proportionally scales removal marginals. Unsupported statements are allocated separately, using
$w(I')=\mathcal V(L\cup I')-\mathcal V(L)$ for $I'\subseteq I$.
Their allocation therefore measures the extra change beyond the supported statements, rather than redistributing the supported set's value. For a set $N$, a value function $f$, and the set $\mathcal P(N)$ of all orders of $N$, the sequential average is
\begin{equation}
    a_\ell=\frac{1}{|N|!}\sum_{\sigma\in\mathcal P(N)}
    \bigl[f(N_{<\ell}^{\sigma}\cup\{\ell\})-f(N_{<\ell}^{\sigma})\bigr],
    \label{eq:shapley-allocation}
\end{equation}
where $N_{<\ell}^{\sigma}$ contains statements before $\ell$ in order $\sigma$. Use $f=\mathcal V$ for the supported set and $f=w$ for the unsupported set.

We convert the allocated values into statement rewards:
\begin{equation}
    r_\ell^{\mathrm{stmt}}=\begin{cases}
        \max(a_\ell,0)-0.01,&\text{supported},\\
        \min(a_\ell,0)-0.01,&\text{unsupported}.
    \end{cases}
    \label{eq:statement-reward}
\end{equation}
The environment separately checks for fabricated rating statements. When a statement is flagged as fabricated, an additional penalty of $0.04$ is applied:
\begin{equation}
r_{\ell}^{\mathrm{stmt}} \leftarrow r_{\ell}^{\mathrm{stmt}} - 0.04\,\mathbf{1}[\text{fabricated}_{\ell}].
\end{equation}
This penalty is based on the environment's verification of the statement and does not enter the Bayesian posterior.

The information term is nonnegative for supported statements and nonpositive for unsupported ones. Every statement pays a reading cost of $0.01$. Repetition can help listeners who missed the rating statement earlier. Otherwise, it usually incurs only that cost. Two additional rules concern truth. A faithful relay of a false rating receives exactly $-0.01$ before response-level clipping. If a speaker corrects its own earlier false rating statement with a true rating, the correction is valued against a record with the earlier false rating statement removed. The scoring baseline therefore excludes the error being corrected.

If the sum of negative statement rewards in one response is below $-2$, those negative rewards are scaled proportionally to sum to $-2$. Positive rewards are unchanged. This bound is applied before token assignment. Statement credit is then set to zero when it is identical across the candidate group. Text without a cell-rating claim has no separate statement score, but may still belong to an action span or receive a retention penalty.

\subsection{Request value and remaining turns}

Suppose a valid request asks holder $q$ for cell $c$, and let $x_{q,c}$ be a truthful rating statement for that cell. We first sum the other agents' surprisal reductions if they all receive this rating statement, including the requester's reduction. Let $p_{\mathrm{next}}(q)$ be the probability that $q$ has another speaking turn. The implemented request reward is
\begin{equation}
\begin{split}
    r_{\mathrm{ask}}(q,c)={}&p_{\mathrm{next}}(q)\\
    &{}\cdot\min\!\left\{2,\,\delta_c\left[\sum_{j\ne q}\bar\Delta_j(x_{q,c})\right]_+\right\},
\end{split}
\label{eq:request-reward}
\end{equation}
where $[z]_+=\max(z,0)$. The factor $\delta_c$ is $0.3$ if the requester has received any rating statement for the cell and $1$ otherwise. The cap, discount, and remaining-turn factor convert the potential information value into a training reward. The calculation assumes a true answer from the holder and receipt by every other present agent. It includes neither a $0.7$ delivery weight nor an estimated willingness to answer. The score is assigned when the request is made and is not revised after observing a reply or a missed message.

For its timing calculation, index seats by $0,\ldots,n-1$ and rounds from zero. Let $s$ be the current speaker, $q\ne s$ the holder, and $\rho$ the current round. The holder's next turn falls in round
\begin{equation}
    r_q=\left\lfloor\frac{n\rho+s+((q-s)\bmod n)}{n}\right\rfloor.
    \label{eq:next-holder-turn}
\end{equation}
Let $T_0=\max(T_{\min},\rho+1)$, using the current family's bounds. Reaching the current round implies $T\ge T_0$. The probability of reaching the holder's next turn is
\begin{equation}
\begin{split}
    p_{\mathrm{next}}(q)&=\Pr(T>r_q\mid T\ge T_0)\\
    &=\frac{[T_{\max}-\max(T_0,r_q+1)+1]_+}{T_{\max}-T_0+1}.
\end{split}
    \label{eq:request-horizon}
\end{equation}
If the holder has yet to speak this round, this probability is one. Otherwise it depends on whether another round occurs. It says nothing about whether the holder will answer.

The requester's prior receipt of any rating statement for the requested cell sets $\delta_c=0.3$, regardless of that rating statement's value. Otherwise $\delta_c=1$. The two-bit cap follows this discount and precedes multiplication by $p_{\mathrm{next}}$. A pending request from the same requester for the same cell sets the information credit to zero. After that request is cleared, a new request is valued from the updated records. If a holder already reported the cell but a listener missed it, a further request can still have value.

\subsection{Action penalties}

A request for a cell already in the requester's briefing, or for a cell with no eligible holder, is invalid and is not registered as pending. It receives no information credit and incurs an action penalty of $0.10$. An answerable request left pending also incurs $0.10$, charged to the requested agent. Let $v_{\mathrm{ask}}(x)$ be the request credit in response $x$, zero if there is no valid credited request. Let $b(x)$ indicate an invalid action and $u_i(x)$ indicate that speaker $i$ leaves an answerable request pending. The action score is
\begin{equation}
\begin{split}
    r^{\mathrm{act}}(x)={}&v_{\mathrm{ask}}(x)-0.05\\
    &{}-0.10\,b(x)-0.10\,u_i(x).
\end{split}
\label{eq:action-score}
\end{equation}
The fixed speaking-turn deduction $0.05$ is shared by every candidate in a group. The action tolerance and subsequent mean subtraction are invariant to this shared constant, so it cancels from all action advantages. It does not discourage message length or speaking through a training gradient. The statement reading cost remains in uncentered statement credit, while the two action penalties can distinguish candidates.

\section{Private-Answer Evaluation}
\label{app:answer-targets}

Let $\hat y_i$ be agent $i$'s selected option, $\hat s_{io}$ its reported total for option $o$, and $s^K_{io}$ the reference total from its available ratings. The implemented private-answer reward is
\begin{equation}
\begin{aligned}
    r_i^{\mathrm{answer}}&=c_i+\frac1k\sum_{o=1}^k\mathbf1[\hat s_{io}=s^K_{io}],\\
    c_i&=\begin{cases}
        \bar P(\hat y_i\mid K_i),&\text{parseable choice},\\
        0,&\text{otherwise}.
    \end{cases}
\end{aligned}
\label{eq:answer_reward}
\end{equation}
The choice term uses the evaluator's posterior for the selected option, floored at $10^{-4}$ when the choice is parseable. An unparseable choice receives zero. No comparison with the true winning label enters this term. A truly incorrect option can therefore earn a high choice score when the current evidence favors it.

The totals term requires one reference value for each cell. Let $v^K_{ic}$ denote the rating used in agent $i$'s known total. Its briefing value takes priority. Otherwise, use the rating statement for cell $c$ whose first-reception time is latest. If no rating statement exists, use zero. The reference total is
\begin{equation}
    s^K_{io}=\sum_{a=1}^m v^K_{i,(o,a)}.
    \label{eq:known-answer-total}
\end{equation}
Each cell contributes once. Missing total lines count as incorrect rather than as zero-valued predictions.

Both posterior computation and the totals target use one retained rating per cell, with initial briefing values taking precedence over received statements. When multiple reports concern the same cell, the most recently received rating is used. An incorrect report may therefore affect an agent's probability estimates and calculated totals until it is corrected. Receiving an old identifier again does not refresh its time. If a true rating statement was followed by a new false rating statement, hearing the old true rating statement again can therefore leave the false value in the totals target. Correctly matching the known totals does not imply recovering the full table's true totals.

For private totals, rules read the number after the final equals sign in a calculation, or the first number after a colon otherwise. If an option has several total lines, its last line supplies the reported total. A missing or unreadable total counts as incorrect. An unreadable choice receives zero choice score.

\section{Self-Play Training Details}
\label{app:training-details}

\subsection{Collection and candidate groups}

Each collection round contains 60 games played by the current shared policy. Sampling uses temperature $1.2$, disabled thinking, and a limit of 1,024 generated tokens per message. Samples containing thinking markers are excluded. A speaking group has 16 policy candidates and up to two additional demonstrations. An intermediate or final private-answer group has eight policy candidates and up to two demonstrations. Demonstrations are separate from the policy sample counts.

One of the 16 policy messages is selected uniformly using the game seed and turn to continue discussion. Reward does not affect this selection. All retained candidates are scored against the same preceding records, so their differences concern the current action rather than different conversation histories. Private answers do not continue the dialogue. Groups with no scoring differences are discarded.

\subsection{Component advantages and token spans}

Statement spans use their own statement reward without a group baseline or standard-deviation normalization. If all candidate statement scores are identical, their statement credit is zeroed. Actions, choices, and each option's total correctness instead use centered advantages. For component $b$ and a group of $G$ policy candidates, define
\begin{equation}
    A_g^b=\tilde r_g^b-\frac1G\sum_{h=1}^G\tilde r_h^b,
    \label{eq:group-advantage}
\end{equation}
where $\tilde r_g^b$ is candidate $g$'s component score, adjusted only for the action tolerance below. Statements are excluded from this centering. Choice and action advantages use this expression directly. Each option's total advantage is additionally scaled by $1/k$.

For actions, let $r_{\max}$ be the maximum among policy candidates and $\tau=0.03$. Before subtracting the policy-group mean, apply
\begin{equation}
    \tilde r_g^{\mathrm{act}}=\begin{cases}
        r_{\max},&r_{\max}-r_g^{\mathrm{act}}<\tau,\\
        r_g^{\mathrm{act}},&\text{otherwise}.
    \end{cases}
    \label{eq:action-tolerance}
\end{equation}
The tolerance treats nearly tied policy actions as equally good. Demonstrations do not enter $r_{\max}$ or the group mean. Choice advantages center the choice scores over eight policy answers. For option $o$, let $\hat s_{go}$ be candidate $g$'s reported total for the current agent $i$, and let $z_{go}=\mathbf1[\hat s_{go}=s^K_{io}]$. Its total-line advantage is
\begin{equation}
    A_{go}^{\mathrm{total}}=\frac1k\left(z_{go}-\frac1G\sum_{h=1}^G z_{ho}\right),
    \qquad G=8.
    \label{eq:total-advantage}
\end{equation}
This compares correctness separately for each option, not only the sum of correct totals.

Every token assigned to a statement receives that statement's credit. Tokens in the request sentence and the end token receive the centered action advantage multiplied by $16/N_{\mathrm{act}}$, where $N_{\mathrm{act}}$ is their token count. This fixes the total action-span weight at 16. Without a request, the closing sentence and end token are used instead. This location is a heuristic for the decision to finish without asking.

For private answers, the choice line and end token receive the choice advantage. Each option's last total line receives its own total advantage. If that line is missing, its advantage is added to the choice span. If no choice line is identifiable, the choice advantage is assigned to answer tokens outside the total lines. Remaining tokens receive zero direct advantage.

Span boundaries use parsed text positions and attribute-word anchors. A token is assigned using the span containing its first character. An action span takes priority over an overlapping statement span. These rules are approximate: for example, a rating word before its attribute can fall in a neighboring span. They associate credit with generated text without guaranteeing perfect attribution at every token.

\subsection{Generated demonstrations}

When the evaluator ranks a request as the best available action, demonstrations provide additional examples of that action. Frozen initial Qwen3-8B minimally rewrites the first usable policy response to express the selected action, with four attempts at temperature $1.0$. Request templates provide two further candidates. The parser filters them, and at most two valid candidates with the lowest negative log likelihood under the current policy are retained. Identical demonstrations within a group are deduplicated.

Demonstrations are trained with the ordinary agent prompt, without rewrite instructions. They are never selected to advance the conversation. Their centered components use the policy group's baseline without changing it. Action demonstration advantages are clipped to $[-0.5,0.5]$. Statement and answer demonstration advantages do not share this cap.

A demonstration was not sampled from the current policy under the ordinary prompt. For the training probability ratio defined below, its denominator is evaluated under the model at the start of the round. The generating model's token probabilities are not used. The resulting objective is a clipped, advantage-weighted imitation update without correction for that generation distribution. It is not an unbiased on-policy PPO estimate. Demonstration tokens receive no reference KL penalty. An entropy term, which encourages a broader token distribution, includes both policy and demonstration tokens.

\subsection{Policy objective and optimization}

Let $\pi_\theta$ be the current model with trainable parameters $\theta$, $\pi_{\mathrm{round}}$ the model at the start of the collection round, and $\pi_0$ the initial backbone before collaboration training. For token $w_t$ with preceding context $h_t$, define the probability ratio and clipped score
\begin{equation}
\begin{split}
    \rho_t(\theta)&=\frac{\pi_\theta(w_t\mid h_t)}{\pi_{\mathrm{round}}(w_t\mid h_t)},\\
    u_t&=\min\{\rho_t A_t,\operatorname{clip}(\rho_t,0.8,1.28)A_t\},\\
    f_t&=\begin{cases}
        \max(u_t,3A_t),&A_t<0,\\
        u_t,&A_t\ge0.
    \end{cases}
\end{split}
\label{eq:policy-objective}
\end{equation}
Here $A_t$ includes the component and span weights above. The extra bound for negative advantages is dual clipping \citep{ye2020moba}. Policy samples use the round's policy as their reference for this ratio. Demonstrations use the same evaluated ratio with the limitation described in the preceding subsection.

For a batch $\mathcal B$ of responses, the policy objective averages responses after dividing each token sum by the fixed constant 64:
\begin{equation}
    J(\theta)=\frac{1}{|\mathcal B|}\sum_{x\in\mathcal B}\frac1{64}\sum_{t\in x}f_t.
    \label{eq:loss-normalization}
\end{equation}
It does not divide by each response's length. Longer statement spans can therefore receive more total gradient weight, whereas action spans have the fixed weight described above. Optimization updates all parameters with AdamW8bit \citep{loshchilov2019adamw,dettmers2022optimizers}, a gradient-norm limit of $1.0$, and 64 samples per step. Each collection round receives one epoch at a constant learning rate, with optimizer state retained across rounds. All three backbones follow the learning-rate schedule in \Cref{tab:learning-rates}.

\subsection{Retaining the initial model's behavior}

A reference penalty applies only to policy-sampled tokens whose assigned advantage is exactly zero. It uses the fixed initial backbone $\pi_0$, not the previous round's model. Let $z_t$ indicate these eligible tokens and $d_t=\log\pi_\theta(w_t\mid h_t)-\log\pi_0(w_t\mid h_t)$. The sampled second-order estimate, commonly called the $k_2$ KL estimate \citep{schulman2020kl}, and the general-prompt imitation loss are
\begin{equation}
\begin{split}
    L_{\mathrm{KL}}&=\mathbb E_t[z_t\,d_t^2/2],\\
    L_{\mathrm{anchor}}&=-\mathbb E_{(x,y)\sim\mathcal D_0}
        \sum_t\log\pi_\theta(y_t\mid x,y_{<t}).
\end{split}
\label{eq:retention-losses}
\end{equation}
Here $\mathcal D_0$ contains general prompts and answers generated by $\pi_0$. The squared log-ratio term is an estimate, not an exact sum over the vocabulary. Both retention terms have a coefficient of $0.1$.

General prompts come from a cleaned SmolTalk subset \citep{allal2025smollm2}, excluding multi-agent prompts and overlap with evaluation data. Approximately 2,500 prompts are used per round, with one anchor sample inserted per four policy samples. Imitating the initial model's answers gives the gradient of a forward KL objective from that model to the current policy. Every ten optimizer steps, a separate set of 170 reference examples monitors this direction of KL. The threshold $0.01$ is checked after training. It does not trigger automatic early stopping.

\subsection{Common configuration}

\Cref{tab:training-common} summarizes settings shared by the training procedure. Qwen3-4B, Qwen3-8B, and Gemma-3-4B-it are trained in separate runs with the same reward definitions and information-tracking rules. The frozen parser uses the initial Qwen3-8B model in all runs. Thinking is disabled for toy-task sampling, training responses, and parsing. These settings describe the training task. \Cref{app:evaluation-details} gives the evaluation settings.

\begin{table}[!ht]
\centering
\small
\begin{tabular}{@{}lr@{}}
\toprule
Setting & Value \\
\midrule
Games per collection round & 60 \\
Policy messages per speaking state & 16 \\
Policy answers per answer state & 8 \\
Additional demonstrations per group & 0--2 \\
Policy sampling temperature & 1.2 \\
Generated message limit & 1,024 tokens \\
Action tolerance & 0.03 \\
Samples per optimizer step & 64 \\
Epochs per collection round & 1 \\
Ratio clipping bounds & $[0.8,1.28]$ \\
Dual clipping bound & 3 \\
Gradient-norm limit & 1.0 \\
Reference KL coefficient & 0.1 \\
General-prompt imitation coefficient & 0.1 \\
\bottomrule
\end{tabular}
\caption{Common sampling and optimization settings. A collection round means generating fresh games and then training on their candidate responses.}
\label{tab:training-common}
\end{table}

\subsection{Learning rates and hardware}

We train each backbone for three collection rounds. \Cref{tab:learning-rates} gives the learning rate for each round. Our experiments use two NVIDIA H100 GPUs and four Intel Gaudi 2 accelerators.

\paragraph{Training time.}
Each collection round, including rollout generation and one optimization epoch, took approximately five hours. The three-round training procedure therefore required approximately 15 hours per complete training run. These estimates cover the CoSIR training procedure only and do not include evaluation or other experimental runs.

\begin{table}[!ht]
\centering\small
\setlength{\tabcolsep}{3pt}
\renewcommand{\arraystretch}{1.12}
\begin{tabular*}{\columnwidth}{@{\extracolsep{\fill}}lccc@{}}
\toprule
Backbone & Round 1 & Round 2 & Round 3 \\
\midrule
Qwen3-8B & $1\times10^{-5}$ & $1\times10^{-5}$ & $2\times10^{-5}$ \\
Qwen3-4B & $1\times10^{-5}$ & $1\times10^{-5}$ & $2\times10^{-5}$ \\
Gemma-3-4B-it & $1\times10^{-5}$ & $1\times10^{-5}$ & $2\times10^{-5}$ \\
\bottomrule
\end{tabular*}
\caption{Learning rates for the three collection rounds. Each rate is fixed within its round.}
\label{tab:learning-rates}
\end{table}

\section{SplitQA Construction and Evaluation}
\label{app:splitqa}

\paragraph{Data selection.} We construct a 70-question SplitQA test set from the distractor setting of HotpotQA \citep{yang2018hotpotqa}, with 63 bridge questions and 7 comparison questions. Each source item contains two annotated supporting Wikipedia paragraphs and eight distractor paragraphs. Bridge questions link information through an intermediate entity, whereas comparison questions compare a property of two entities. We retain only bridge and comparison questions with two distinct supporting paragraphs. We also exclude any item whose gold answer appears verbatim in a distractor paragraph. This removes direct copies of the answer from the distractors.

\paragraph{Evidence allocation.} Agent~1 receives one supporting paragraph and Agent~2 receives the other. Agents~3 and~4 receive only distractor paragraphs. Thus, no agent receives both annotated supporting paragraphs, and exactly two agents hold the annotated evidence. To answer from that evidence, the team needs to identify which agents hold relevant facts and link or compare those facts across the two supporting paragraphs. The distractors add the need to separate relevant information from unrelated material.

\paragraph{Discussion and scoring.} We retain the protocol used in our HiddenBench evaluation \citep{li2026hiddenbench}: four agents speak in a fixed order for five rounds, then each answers privately. Each round gives every agent one speaking turn. Final answers are free-form strings, so agents are not given a list of candidate answers. The team answer is the unique answer receiving the most votes among the four agents. If two or more answers tie for the highest vote count, the team produces no answer and receives an exact-match score of zero, even if one of the tied answers is correct. Otherwise, the team receives a score of one if the selected answer exactly matches the gold answer, and zero if it does not. Each of the 70 questions is run with four seeds.

\section{Evaluation Settings and Score Aggregation}
\label{app:evaluation-details}

\paragraph{Benchmark protocols.}
HiddenBench \citep{li2026hiddenbench} uses five rounds of discussion in a fixed speaking order, followed by private votes from all agents. Team accuracy is determined by majority voting. SILO-BENCH \citep{zhang2026silo} uses three discussion rounds followed by individual answers, and we report its task success rate. SplitQA uses five rounds followed by private answers, with team accuracy measured by exact match of the answer receiving the most votes. Tied votes are counted as incorrect, as detailed in \Cref{app:splitqa}.

MathChat \citep{wu2023mathchat} pairs an LLM with a user proxy that executes tools and provides feedback during mathematical problem solving. BenchForm \citep{weng2025conformity} requires a single answer rather than a new multi-round discussion. We average answer accuracy across its five settings of peer influence. In the trust and doubt settings, the input includes five rounds of prior discussion history. For Collaboration Tax \citep{sun2026collaboration}, we report the score achieved by the collaborating agents. Higher scores therefore indicate better performance, as they do for the other tasks in SUM.

\paragraph{Generation settings.}
\Cref{tab:evaluation-decoding} lists the temperature, output limit, and thinking setting for each task. The token limit is the maximum number of tokens generated in one response. Thinking refers to the model's thinking mode, where supported. HiddenBench, SILO-BENCH, and SplitQA use top-$p=0.95$ during discussion. Their final answers use temperature zero with thinking enabled.

For SILO-BENCH, if an agent produces no final answer, we ask it to answer once more with thinking disabled. In Einstein Puzzles \citep{peng2025communication}, the sampling seed for each step is determined by the instance index and step index.

\begin{table*}[t]
\centering\small
\setlength{\tabcolsep}{6pt}
\renewcommand{\arraystretch}{1.12}
\begin{tabular*}{\textwidth}{@{\extracolsep{\fill}}llrrc@{}}
\toprule
Task & Response stage & Temperature & Token limit & Thinking \\
\midrule
HiddenBench & Discussion & 0.7 & 320 & Off \\
& Final answer & 0 & 10,240 & On \\
\addlinespace[3pt]
SILO-BENCH & Discussion & 0.7 & 320 & Off \\
& Final answer & 0 & 6,144 & On \\
\addlinespace[3pt]
SplitQA & Discussion & 0.7 & 320 & Off \\
& Final answer & 0 & 6,144 & On \\
\midrule
MathChat & Interaction & 0.2 & 1,024 & Off \\
BenchForm & Answer & 0.2 & 600 & Off \\
Einstein Puzzles & Step & 0.2 & 600 & Off \\
Collaboration Tax & Interaction & 0.7 & 4,096 & Off \\
\bottomrule
\end{tabular*}
\caption{Generation settings for the seven unseen collaboration tasks. The first three tasks use separate settings for discussion and final answers. SILO-BENCH permits one further answer attempt with thinking disabled when needed, as described in the text.}
\label{tab:evaluation-decoding}
\end{table*}

\paragraph{Inference-time collaboration workflows.}
ToM \citep{li2023theory} and TalkHier \citep{wang2025talk} follow the same benchmark protocols as the other methods, with matched numbers of agents, discussion rounds, and limits on speaking turns. However, they use additional model calls within each speaking turn. ToM makes two calls. The first generates a private belief-state description of at most 400 tokens, which is retained in the agent's context but not shared with teammates. The second produces the public message. TalkHier makes three calls. It first drafts a response, then obtains reviewer feedback of at most 400 tokens, and finally revises the response using that feedback. These additional calls do not increase the number of public speaking turns, but they increase inference computation.

\paragraph{Score aggregation.}
\Cref{tab:ood-results} reports means over the runs available for each task. All OOD benchmarks have four reported runs per method and model. We first average within each task, then add the seven means to obtain SUM. We compute the total before rounding the displayed entries to two decimal places. Consequently, adding the rounded table entries can differ slightly from the displayed total.

Each task contributes equally to SUM. A one-point increase on any task adds one point to the total, regardless of that task's number of examples or runs. The general-capability benchmarks and behavioral measurements are not included in SUM. The column $\Delta$SUM subtracts the corresponding Base total, so it measures the change from the same starting model. Bold entries identify the highest score within a model and task, including ties.

The comparison methods serve different purposes. ToM and TalkHier modify the inference-time workflow, whereas Outcome-RL uses CoSIR's training procedure with the reward replaced by the final team outcome. Base measures performance before collaboration training. The Outcome-RL comparison therefore isolates the role of behavior-specific training rewards (Unlike CoSIR's candidate-level information scoring, Outcome-RL assigns final-outcome feedback to the sampled trajectory. Both methods retain request demonstrations), while the inference-time methods test whether collaboration can instead be improved through additional reasoning and revision steps. 

\section{General Capability Retention}
\label{app:general-capabilities}

We compare Base and CoSIR on mathematical reasoning with GSM8K \citep{cobbe2021verifiers}, general knowledge with MMLU \citep{hendrycks2021mmlu}, and instruction following with IFEval \citep{zhou2023ifeval}. \Cref{tab:general-capabilities} gives the scores for each model. Every change lies between $-0.6$ and $+0.8$ percentage points, so performance remains close to the starting checkpoints on these benchmarks.

\begin{table}[t]
\centering\small
\setlength{\tabcolsep}{3.5pt}
\renewcommand{\arraystretch}{1.06}
\begin{tabular*}{\columnwidth}{@{\extracolsep{\fill}}llrrr@{}}
\toprule
Model & Policy & GSM8K & MMLU & IFEval \\
\midrule
Qwen3-8B & Base & 93.9 & 77.6 & 84.4 \\
& CoSIR & 93.7 & 77.8 & 84.1 \\
\addlinespace[3pt]
Qwen3-4B & Base & 93.3 & 75.3 & 81.5 \\
& CoSIR & 93.1 & 74.7 & 81.8 \\
\addlinespace[3pt]
Gemma-3-4B-it & Base & 88.9 & 63.5 & 75.1 \\
& CoSIR & 89.7 & 63.8 & 75.0 \\
\bottomrule
\end{tabular*}
\caption{General-capability scores (\%) before and after collaboration training.}
\label{tab:general-capabilities}
\end{table}

\section{Behavioral Measures}
\label{app:behavior-details}

The measures in \Cref{tab:behavior-results} track observable actions in the discussion. Their denominators differ because some concern information holders and others concern messages. All are reported as percentages.

\paragraph{HiddenBench.}
\textit{Private facts disclosed} is the fraction of private facts stated by their original holders during the discussion. The disclosure check matches a fact when at least 60\% of its content words appear in the holder's message. \textit{Messages responding to teammates} is the fraction of messages that respond to another agent's contribution. This measure tracks engagement with a teammate without requiring that the response be correct.

\paragraph{SILO-BENCH.}
\textit{Agents listing all own data} is the fraction of agents that list their complete local data in their messages. \textit{Result-only messages} is the fraction of messages that relay an accumulated value or conclusion without giving the speaker's own data. The first measure increases when agents expose inputs that teammates can check. The second decreases when fewer messages offer a result without those inputs.

\paragraph{SplitQA.}
\textit{Holders stating supporting facts} is the fraction of agents assigned a supporting paragraph that state their supporting fact in at least one discussion round. Only the two supporting-fact holders enter the denominator. Agents assigned only distractors are excluded.

\paragraph{MathChat.}
\textit{Messages with correction language} is the fraction of discussion messages containing language that identifies or corrects an error. It measures attempts to correct the discussion, without separately scoring whether each correction is valid.

\paragraph{Timing of disclosure.}
The case-study analysis also measures whether evidence is shared in the holder's first message. For Qwen3-8B, this rate rises from 61.1\% to 69.0\% on HiddenBench and from 41.1\% to 51.1\% on SplitQA. These first-message rates are distinct from the any-round disclosure rates in \Cref{tab:behavior-results}. They indicate that more relevant evidence enters the discussion at its start, when teammates can use it in subsequent turns.

\section{MAST Failure Analysis}
\label{app:mast-results}

We use the MAST categories of \citet{cemri2025fail} to organize the observed failures. \Cref{tab:mast-results} pools conversations from HiddenBench, SplitQA, and MathChat across Qwen3-8B, Qwen3-4B, and Gemma-3-4B-it. A category rate is the percentage of conversations marked with that category. A conversation can have several failure labels, so the percentages do not sum to 100.

We follow the official MAST evaluation procedure \citep{cemri2025fail}, using the original judge prompt and the definitions and examples of all 14 failure modes without modification. We replace only the judge model with gpt-5-mini, using medium reasoning effort and a maximum output of 8,000 tokens. We apply this procedure to Base and CoSIR conversations from HiddenBench, SplitQA, and MathChat for all three backbones: Qwen3-8B, Qwen3-4B, and Gemma-3-4B-it. A conversation may receive multiple failure labels, so the reported percentages do not need to sum to 100.

Failures involving coordination between agents decrease from 67.8\% to 62.0\%, and failures involving verification or termination decrease from 58.8\% to 53.5\%. The specification and system-design category remains nearly unchanged, at 88.2\% and 88.3\%. Within these categories, ignoring teammates' input and proceeding without clarification become less frequent, while information withholding changes from 18.5\% to 18.7\%. Thus, the aggregate reduction concerns particular ways of using information, rather than every failure related to communication.

The disclosure measures count whether designated private facts are shared, whereas MAST information withholding is a conversation-level failure judgment. The two measures therefore capture different aspects of information sharing.

\section{Case Studies}
\label{app:case-studies}

The following cases compare the initial Qwen3-8B with its trained checkpoint after the third training round. Each pair uses the same item and seed, without an additional situational prompt. Both conditions use four agents. HiddenBench and SplitQA allow five discussion rounds, and the SILO-BENCH example allows three. A round gives each agent one speaking turn. The cards reproduce selected excerpts, with omissions marked by an ellipsis. Descriptions outside quotation marks summarize the supplied dialogues.

\subsection{HiddenBench: Sharing Facts That Change a Decision}

The team must select a warehouse for relief supplies after a storm (seed 3). Shared information makes Central Depot attractive because it is near a highway and has modern loading facilities. The private evidence changes this comparison. Agent~1 knows that Bayview Warehouse has lost half its roof and cannot keep supplies dry. Agent~2 knows that flooding has put Central Depot's foundation at risk. Agent~3 knows that Central Depot's backup generator has been destroyed, and Agent~4 knows that Hilltop Storage is the only site with a full staff. The correct choice is Hilltop Storage.

Agent~1 initially recommends Central Depot in both conditions. In Base, Agents~3 and~4 never state their private facts, and all four agents ultimately select Bayview Warehouse. After training, both facts appear in the first round. Agent~1 takes up the generator failure in round 2 and recommends Hilltop Storage by round 5. All four agents select it. \Cref{fig:case-hiddenbench} shows the exchange. The change is visible in the discussion itself: an initial recommendation is revised after teammates provide evidence that was absent from the shared briefing.

\begin{figure*}[t]
\centering
\begingroup\setlength{\fboxsep}{8pt}
\fcolorbox{black!20}{black!2}{\begin{minipage}{\dimexpr\textwidth-2\fboxsep-2\fboxrule\relax}
\small
\textbf{HiddenBench: new evidence changes the initial recommendation}\par\medskip
\begin{minipage}[t]{0.475\linewidth}\raggedright
\textbf{Base}\par\smallskip
\textbf{Agent 3, round 1:}\par
``While the foundation issue at Central Depot is concerning, the briefing does not provide any evidence that the warehouse is currently unsafe or structurally compromised.\ \ldots\ If we choose a warehouse with a damaged generator, we risk delays in power supply, which could impact refrigeration or lighting for the supplies.''\par\smallskip
\textbf{Agent 4, round 5:}\par
``Given the urgency of the situation, the need for a stable and secure hub, and the risk of structural failure at Central Depot, I fully support the recommendation of Bayview Warehouse.''\par\smallskip
\textbf{Final answers:} Bayview Warehouse, all four agents (incorrect).
\end{minipage}\hfill
\begin{minipage}[t]{0.475\linewidth}\raggedright
\textbf{CoSIR}\par\smallskip
\textbf{Agent 3, round 1:}\par
``Now, regarding Central Depot, the backup generator was destroyed by lightning and cannot be fixed for several days.''\par\smallskip
\textbf{Agent 4, round 1:}\par
``Hilltop Storage, on the other hand, did not lose any personnel and has a full staff to unload and pack supplies.''\par\smallskip
\textbf{Agent 1, round 2:}\par
``It's clear now that Central Depot's backup generator is completely out of commission, which means it can't support operations for several days.''\par\smallskip
\textbf{Agent 1, round 5:}\par
``Therefore, I recommend Hilltop Storage as the main distribution hub.''\par\smallskip
\textbf{Final answers:} Hilltop Storage, all four agents (correct).
\end{minipage}
\end{minipage}}
\endgroup
\caption{HiddenBench warehouse selection, seed 3. Both teams begin with Agent 1 recommending Central Depot. After training, the missing facts enter the discussion and Agent 1 revises its choice.}
\label{fig:case-hiddenbench}
\end{figure*}

\subsection{SILO-BENCH: Making a Wrong Count Checkable}

The team must count occurrences of \texttt{delta} across 16 words divided equally among four agents (seed 1). Their word lists contain two, one, two, and zero occurrences, respectively, so the correct total is five. In Base, Agent~2 incorrectly reports two occurrences in its own list. Agents repeat their counts without exposing their word lists, and all four finish with the incorrect total of 18.

The trained team also makes an initial counting error: Agent~3 reports zero despite holding two occurrences. In round 2, it lists its words while still asserting the incorrect count. Agent~1 then checks that list and corrects the contribution to two, obtaining $2+1+2+0=5$. All four agents finish with the correct answer. \Cref{fig:case-silo} shows that sharing the underlying data makes a local error visible to a teammate who can repair it.

\begin{figure*}[t]
\centering
\begingroup\setlength{\fboxsep}{8pt}
\fcolorbox{black!20}{black!2}{\begin{minipage}{\dimexpr\textwidth-2\fboxsep-2\fboxrule\relax}
\small
\textbf{SILO-BENCH: a teammate can correct a count once the data are visible}\par\medskip
\begin{minipage}[t]{0.475\linewidth}\raggedright
\textbf{Base}\par\smallskip
\textbf{Agent 2, round 1:}\par
``I count 2 occurrences of 'delta' in my data.''\par\smallskip
\textbf{Agent 2, round 3:}\par
``[Round 3] Agent 2: I count 2 occurrences of 'delta' in my data.''\par\smallskip
Agent 2 actually holds only one occurrence. The local counts are repeated without the raw word lists.\par\smallskip
\textbf{Final answers:} 18, all four agents (incorrect).
\end{minipage}\hfill
\begin{minipage}[t]{0.475\linewidth}\raggedright
\textbf{CoSIR}\par\smallskip
\textbf{Agent 3, round 2:}\par
``[Round 2] Agent 3: I confirm my count of 'delta' is 0. I will now share my data: delta, delta, beta, epsilon.''\par\smallskip
\textbf{Agent 1, round 3:}\par
``[Round 3] Agent 1: Agent 3 shared: delta, delta, beta, epsilon.\ \ldots\ Agent 3's data: 2 'delta's\ \ldots\ Total: 2 + 1 + 2 + 0 = 5''\par\smallskip
Agent 3 still gives a wrong local count, but Agent 1 checks the shared list and corrects it.\par\smallskip
\textbf{Final answers:} 5, all four agents (correct).
\end{minipage}
\end{minipage}}
\endgroup
\caption{SILO-BENCH word counting, seed 1. The trained team succeeds despite an initial local error because the raw data become available for checking.}
\label{fig:case-silo}
\end{figure*}

\subsection{SplitQA: Connecting Facts Held by Different Agents}

The question asks which Eminem album included vocals from a singer who had an album titled \textit{Unapologetic}. We use seed 0 for HotpotQA. Agent~1's paragraph identifies Rihanna as the singer associated with \textit{Unapologetic}. Agent~2's paragraph states that Eminem's song \textit{The Monster} features Rihanna and appears on \textit{The Marshall Mathers LP 2}. Combining these facts yields the answer.

In Base, Agent~2 does not state its supporting fact. Instead, it asserts an unsupported connection to \textit{Recovery}, which becomes all four agents' final answer. After training, Agent~1 gives its evidence and identifies the missing link: which Eminem album includes Rihanna's vocals. Agent~2 supplies that link in the same round. Agent~1 combines the two facts in round 2, and all four agents answer \textit{The Marshall Mathers LP 2}. \Cref{fig:case-splitqa} shows how stating both known and missing information helps the team connect the evidence.

\begin{figure*}[t]
\centering
\begingroup\setlength{\fboxsep}{8pt}
\fcolorbox{black!20}{black!2}{\begin{minipage}{\dimexpr\textwidth-2\fboxsep-2\fboxrule\relax}
\small
\textbf{SplitQA: identify the missing link, then combine the two supporting facts}\par\medskip
\begin{minipage}[t]{0.475\linewidth}\raggedright
\textbf{Base}\par\smallskip
\textbf{Agent 2, round 1:}\par
``I need to clarify the connection between the album `Unapologetic' and Eminem. `Unapologetic' is an album by Rihanna, and while Rihanna did collaborate with Eminem on the song `Numb' from his album `Recovery,' that doesn't mean `Unapologetic' includes Eminem's vocals.''\par\smallskip
The holder does not share the provided fact about \textit{The Monster}. Its claim about \textit{Recovery} has no support in the assigned paragraph.\par\smallskip
\textbf{Final answers:} \textit{Recovery}, all four agents (incorrect).
\end{minipage}\hfill
\begin{minipage}[t]{0.475\linewidth}\raggedright
\textbf{CoSIR}\par\smallskip
\textbf{Agent 1, round 1:}\par
``However, I don't have information about which of Eminem's albums included vocals from Rihanna. I'll need to rely on other team members' input to confirm this.''\par\smallskip
\textbf{Agent 2, round 1:}\par
``From my background material, I know that `The Monster' is a song by Eminem featuring Rihanna, and it is from his album The Marshall Mathers LP 2 (2013).''\par\smallskip
In round 2, Agent 1 connects Rihanna's \textit{Unapologetic} to the Eminem album named by Agent 2.\par\smallskip
\textbf{Final answers:} \textit{The Marshall Mathers LP 2}, all four agents (correct).
\end{minipage}
\end{minipage}}
\endgroup
\caption{SplitQA, seed 0. The correct answer follows from combining the facts held by Agents 1 and 2.}
\label{fig:case-splitqa}
\end{figure*}

\clearpage
\end{document}